\documentclass[10pt,twocolumn,letterpaper]{article}

\usepackage{cvpr}
\usepackage{times}
\usepackage{epsfig}
\usepackage{graphicx}
\usepackage{amsmath}
\usepackage{amssymb}
\usepackage{subfigure}
\def\ie{{\em i.e.}}
\def\eg{{\em e.g.}}
\def\etal{{\em et al.}}

\usepackage[breaklinks=true,bookmarks=false]{hyperref}

\cvprfinalcopy

\def\ie{{\em i.e.}}
\def\eg{{\em e.g.}}
\def\etal{{\em et al.}}

\def\cvprPaperID{****}

\ifcvprfinal\fi
\begin{document}

\title{Improved Selective Refinement Network for Face Detection}

\author{Shifeng Zhang$^{1*}$, Rui Zhu$^{2*}$, Xiaobo Wang$^{2}$\thanks{These authors contributed equally to this work.}, Hailin Shi$^{2}$\thanks{Corresponding author}, Tianyu Fu$^{2}$, Shuo Wang$^{2}$, Tao Mei$^{2}$, Stan Z. Li$^{1}$\\
$^1$ CBSR \& NLPR, Institute of Automation, Chinese Academy of Sciences, Beijing, China. \\
$^2$ JD AI Research, Beijing, China. \\
{\tt\small \{shifeng.zhang,szli\}@nlpr.ia.ac.cn} \\{\tt \small \{zhurui10,wangxiaobo8,shihailin,futianyu,wangshuo30,tmei\}@jd.com}
}

\maketitle
\thispagestyle{empty}

\begin{abstract}
As a long-standing problem in computer vision, face detection has attracted much attention in recent decades for its practical applications. With the availability of face detection benchmark WIDER FACE dataset, much of the progresses have been made by various algorithms in recent years. Among them, the Selective Refinement Network (SRN) face detector introduces the two-step classification and regression operations selectively into an anchor-based face detector to reduce false positives and improve location accuracy simultaneously. Moreover, it designs a receptive field enhancement block to provide more diverse receptive field. In this report, to further improve the performance of SRN, we exploit some existing techniques via extensive experiments, including new data augmentation strategy, improved backbone network, MS COCO pretraining, decoupled classification module, segmentation branch and Squeeze-and-Excitation block. Some of these techniques bring performance improvements, while few of them do not well adapt to our baseline. As a consequence, we present an improved SRN face detector by combining these useful techniques together and obtain the best performance on widely used face detection benchmark WIDER FACE dataset. 
\end{abstract}

\let\thefootnote\relax\footnotetext{This work was mainly done at JD AI Research and supported by JDGrapevine Plan.}

\section{Introduction}
Face detection is the primary procedure for other face-related tasks including face alignment, face recognition, face animation, face attribute analysis and human computer interaction, to name a few. The accuracy of face detection systems has a direct impact on these tasks, hence the success of face detection is of crucial importance. Given an arbitrary image, the goal of face detection is to determine whether there are any faces in the image, and if present, return the image location and extent of each face. In recent years, great progress has been made on face detection~\cite{DBLP:conf/cvpr/LiLSBH15, DBLP:conf/cvpr/QinYLH16, DBLP:conf/eccv/LiSWW16, DBLP:conf/mm/YuJWCH16, hao2017scale, song2018beyond, shi2018real, bai2018finding, zhang2018detecting, zhang2017faceboxes} due to the development of deep convolutional neural network (CNNs)~\cite{vgg, res, googlev1, densely} and the collection of WIDER FACE benchmark dataset~\cite{yang2016wider}. This challenging dataset has a high degree of variability in scale, pose and occlusion as well as plenty of tiny faces in various complex scenes, motivating a number of robust CNN-based algorithms.

\begin{figure*}
\centering
\includegraphics[width=0.75\textwidth]{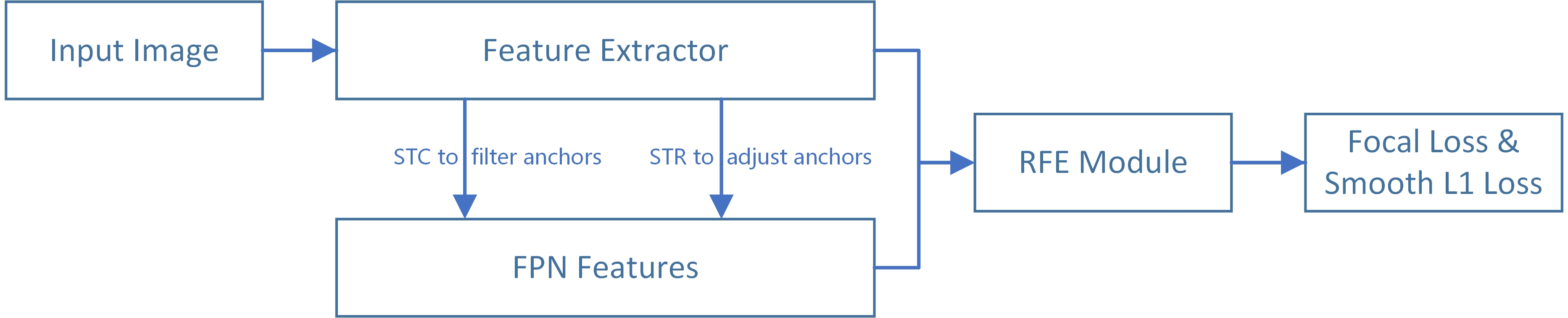}
\caption{The brief overview of Selective Refinement Network. It consists of Selective Two-step Classification (STC), Selective Two-step Regression (STR) and Receptive Field Enhancement (RFE).}
\vspace{-5mm}
\label{fig:structure}
\end{figure*}

We first give a brief introduction to these algorithms on the WIDER FACE dataset as follows. ACF~\cite{yang2014aggregate} borrows the concept of channel features to the face detection domain. Faceness~\cite{yang2015facial} formulates face detection as scoring facial parts responses to detect faces under severe occlusion. MTCNN~\cite{zhang2016joint} proposes a joint face detection and alignment method using unified cascaded CNNs for multi-task learning. CMS-RCNN~\cite{zhu2016cms} integrates contextual reasoning into the Faster R-CNN algorithm to help reduce the overall detection errors. LDCF+~\cite{ohn2016boost} utilizes the boosted decision tree classifier to detect faces. The face detection model for finding tiny faces~\cite{hu2017finding} trains separate detectors for different scales. Face R-CNN~\cite{wang2017face} and Face R-FCN~\cite{wang2017detecting} apply Faster R-CNN~\cite{faster} and R-FCN~\cite{RFCN} in face detection and achieve promising results. ScaleFace~\cite{yang2017face} detects different scales of faces with a specialized set of deep convolutional networks with different structures. SSH~\cite{najibi2017ssh} adds large filters on each prediction head to merge the context information. SFD~\cite{zhang2017s} compensates anchors for small faces with a few strategies in SSD~\cite{ssd} framework. MSCNN~\cite{cai2016unified} performs detection at multiple output layers so as to let receptive fields match objects of different scales. Based on RetinaNet~\cite{focal}, FAN~\cite{wang2017fan} proposes an attention mechanism at anchor level to detect the occluded faces. Zhu \etal~\cite{zhu2018seeing} propose an Expected Max Overlapping score to evaluate the quality of anchor matching. PyramidBox~\cite{tang2018pyramidbox} takes advantage of the information around human faces to improve detection performance. FDNet~\cite{zhang2018face} employs several training and testing techniques to Faster R-CNN to perform face detection. Inspired by RefineDet~\cite{zhang2018single}, SRN~\cite{chi2018selective} appends another binary classification and regression stage in RetinaNet, in order to filter out most of simple negative anchors in the large feature maps and coarsely adjust the locations of anchors in the high level feature maps. FANet~\cite{zhang2017feature} aggregates higher-level features to augment lower-level features at marginal extra computation cost. DSFD~\cite{li2018dsfd} strengthens the representation ability by a feature enhance module. DFS~\cite{tian2018learning} introduces a more effective feature fusion pyramid and a more efficient segmentation branch to handle hard faces. VIM-FD~\cite{zhang2019robust} combines many previous techniques on SRN and achieves the state-of-the-art performance.

In this report, we exploit some existing techniques from classification and detection tasks to further improve the performance of SRN, including data augmentation strategy, improved backbone network, MS COCO pretraining, decoupled classification module, segmentation branch and SE block. By conducting extensive experiments, we share some useful techniques that make SRN regain the state-of-the-art performance on WIDER FACE. Meanwhile, we list some techniques that do not work well in our model, probably because (1) we have a strong baseline that causes them to not work well, (2) combination of ideas is not trivial, (3) they are not robust enough for universality, and (4) our implementation is wrong. This does not mean that they are not applicable to other models or other datasets.

\section{Review of Baseline}
In this section, we present a simple review of our baseline Selective Refinement Network (SRN). As illustrated in Figure~\ref{fig:structure}, it consists of the Selective Two-step Classification (STC), Selective Two-step Regression (STR) and Receptive Field Enhancement (RFE). These three module are elaborated as follows.

\subsection{Selective Two-step Classification}
For one-stage detectors, numerous anchors with extreme positive/negative sample ratio (\eg, there are about $300k$ anchors and the positive/negative ratio is approximately $0.006\%$ in SRN) leads to quite a few false positives. Hence it needs another stage like RPN to filter out some negative examples. Selective Two-step Classification, inherited from RefineDet, effectively rejects lots of negative anchors and alleviates the class imbalance problem. 

Specifically, most of anchors (\ie, $88.9\%$) are tiled on the first three low level feature maps, which do not contain adequate context information. So it is necessary to apply STC on these three low level features. Other three high level feature maps only produce $11.1\%$ anchors with abundant semantic information, which is not suitable for STC. To sum up, the application of STC on three low level features brings advanced results, while on three high level ones will bring ineffective results and more computational cost. STC module suppresses the amount of negative anchors by a large margin, leading the positive/negative sample ratio about $38$ times increased (\ie, from around $1$:$15441$ to $1$:$404$). The shared classification convolution module and the same binary Focal Loss are used in the two-step classification, since both of the targets are distinguishing the faces from the background. 

\subsection{Selective Two-step Regression}
Multi-step regression like Cascade RCNN~\cite{cai2018cascade} can improve the accuracy of bounding box locations, especially in some challenging scenes, \eg, MS COCO-style evaluation metrics. However, applying multi-step regression to the face detection task without careful consideration may hurt the detection results. 

For SRN, the numerous small anchors from three low level feature maps will cause the loss to bias towards regression problem and hinder the essential classification problem. Meanwhile, the feature representations of three lower pyramid levels for small faces are coarse, leading to the obstacle to perform two-step regression. These concerns will not happen while preforming two-step regression on the three high level features, whose detailed features of large faces with large anchor scales help regress to more accurate locations. In summary, Selective Two-step Classification and Regression is a specific and efficient variant of RefineDet on face detection task, especially for small faces and some false positives.

\subsection{Receptive Field Enhancement}
Current networks usually possess square receptive fields, which affect the detection of objects with different aspect ratios. To address this issue, SRN designs a Receptive Field Enhancement (RFE) to diversify the receptive field of features before predicting classes and locations, which helps to capture faces well in some extreme poses.

\section{Description of Improvement}
Here we share some existing techniques that make SRN regain the state-of-the-art performance on the WIDER FACE dataset, including data augmentation, feature extractor and training strategy.   

\subsection{Data Augmentation}
We use the original data augmentation strategies of SRN including photometric distortions, randomly expanding by zero-padding operation, randomly cropping patches from images and resizing patches to $1024\times1024$. Additionally, with probability of $1/2$, we utilize the data-anchor-sampling in PyramidBox~\cite{tang2018pyramidbox}, which randomly selects a face in an image and crops sub-image based anchor. These data augmentation methods are crucial to prevent over-fitting and construct a robust model.

\subsection{Feature Extractor}
The greatest challenge in WIDER FACE is to accurately detect plenty of tiny faces. We believe that the ResNet-50-FPN~\cite{fpn} backbone of SRN still remains considerable room to improve the accuracy, especially for the tiny faces. Root-ResNet from ScratchDet~\cite{zhu2018scratchdet} aims to improve the detection performance of small object, but its training speed is much slower than ResNet. To balance training efficiency and detection accuracy, we improve the ResNet-50 by taking the advantages of Root-ResNet and DRN~\cite{yu2017dilated}.

\begin{figure}[!h]
\centering
\includegraphics[width=0.465\textwidth]{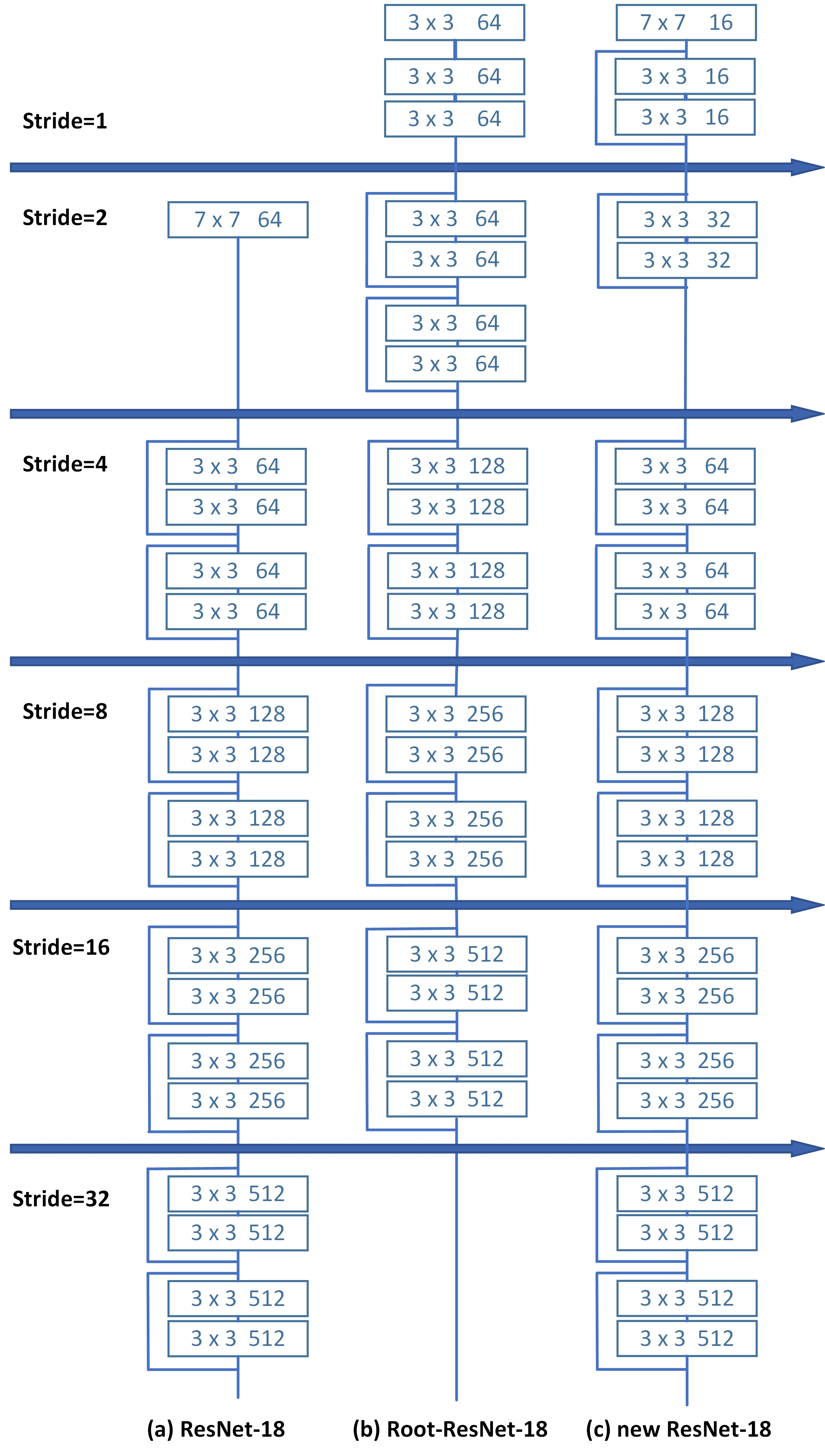}
\caption{Network structure illustration. (a) ResNet-18: original structure. (b) Root-ResNet-18: replacing the $7\times7$ conv layer with three stacked $3\times3$ conv layers and changing the stride $2$ to $1$. (c) New-ResNet-18: combining DRN with Root-ResNet-18 to have a training speed/accuracy trade-off backbone for SRN.}
\label{fig:jiegou0}
\end{figure}

Specifically, the downsampling operation (stride=$2$) to the image in the first $7\times7$ convolution layer of ResNet will cause the loss of important information, especially for small faces. After considering the motivation of Root-ResNet and DRN, we change the first conv layer's stride from $2$ to $1$ and channel number from $64$ to $16$, as well as add two residual blocks (see Figure \ref{fig:jiegou0}). One residual block is for enriching representational information while the other is for downsampling, whose channel number are reduced to $16$ and $32$ to balance the parameters. This configuration can keep essential information of small faces without additional overhead.

\begin{figure*}[t]
\centering
\includegraphics[width=1.0\textwidth]{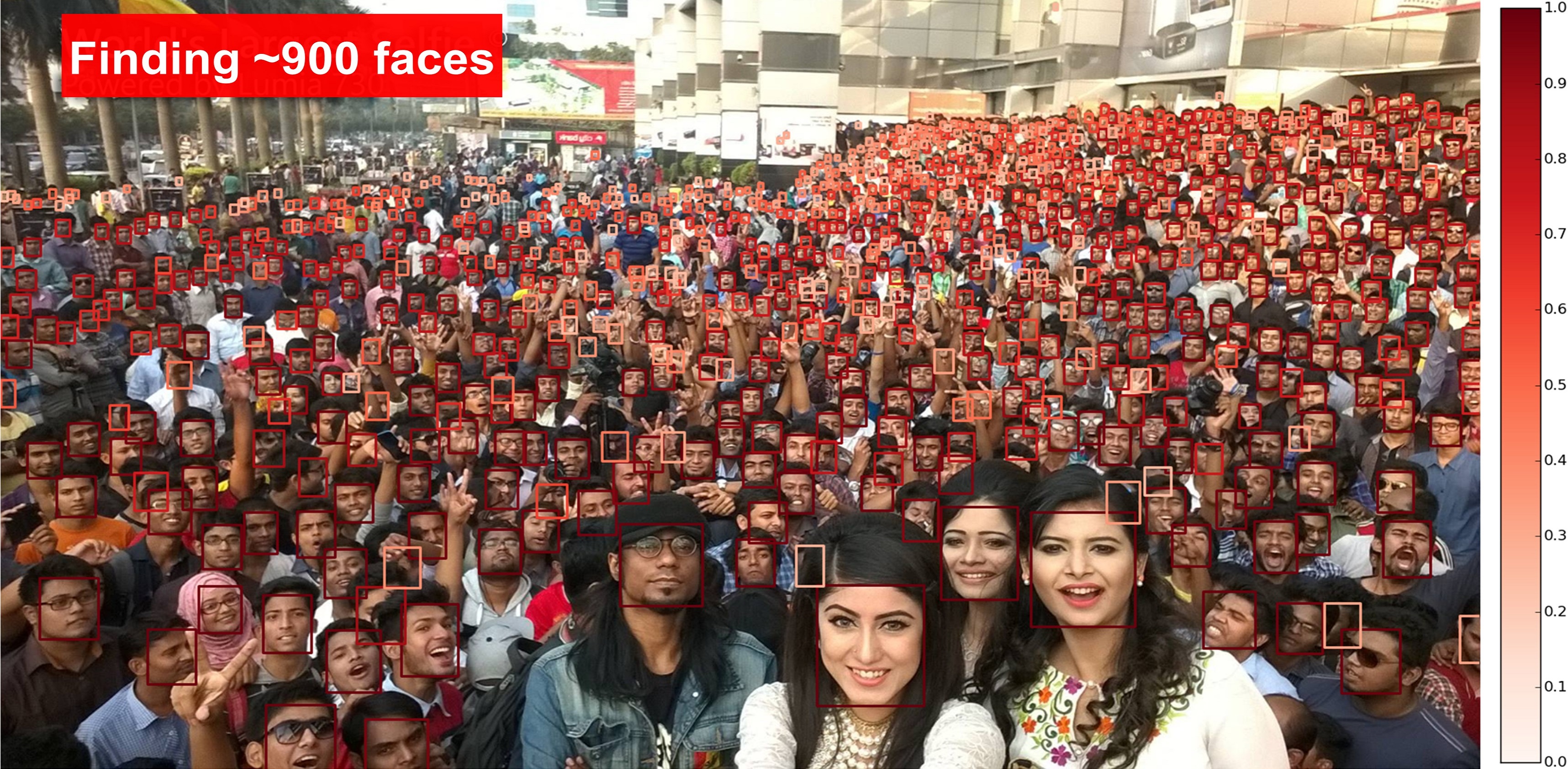}
\caption{A qualitative result. Our detector successfully finds about $900$ faces out of the reported $1000$ faces in the above image. The confidences of the detections are presented in the color bar on the right hand. Best view in color.}
\label{fig:slumia}
\end{figure*}

\subsection{Training Strategy}
Because our ResNet-50-FPN backbone have been modified, we can not use the ImageNet pretrained model. One solution is like DRN that trains the modified backbone on ImageNet dataset~\cite{imagenet} and then finetunes on WIDER FACE. However, He \etal~\cite{he2018rethinking} and ScratchDet have proved that the ImageNet pretraining is not necessary. Thus, we double the training epoch to $260$ epochs and train the model with modified backbone from scratch. One of the key factor to train from scratch is the normalization. Due to the large input size (\ie, $1024\times1024$), one $24$G GPU only can be input up to $5$ images, causing Batch Normalization~\cite{batchnorm} to not work well during training from scratch. To this end, we utilize Group Normalization~\cite{wu2018group} with group=$16$ to train this modified ResNet-50 backbone from scratch. 

Besides, recent work FA-RPN~\cite{najibi2018fa} demonstrates that pretraining the model on the MS COCO dataset~\cite{lin2014microsoft} is helpful to improve the performance of face detector on the WIDER FACE dataset. We attribute this promotion to a number of examples from people category and the objects with similar small scale (\ie, ground truth area $<32$) in the MS COCO dataset. So we also apply this pretraining strategy.

\subsection{Implementation Detail}
{\noindent \textbf{Anchor Setting and Matching}.} Two anchor scales (\ie, $2S$ and $2\sqrt{2S}$, where $S$ represents the total stride size at each pyramid level) and one aspect ratios (\ie, $1.25$) cover the input images (\ie, $1024\times1024$), with the anchor scale ranging from $8$ to $362$ pixels across pyramid levels. We assign anchors with IOU $> \theta_{p}$ as positive, anchors with IOU in $[0, \theta_{n})$ as negative and others as ignored examples. Empirically, we set $\theta_{n}=0.3$ and $\theta_{p}=0.7$ for the first step, and $\theta_{n}=0.4$ and $\theta_{p}=0.5$ for the second step. 

{\noindent \textbf{Optimization}.} At the training process, we simply sum the STC loss and the STR loss. We pretrain the new-designed backbone network with GroupNorm on MS COCO and finetune on WIDER FACE training set using SGD with $0.9$ momentum, $0.0001$ weight decay, and batch size $32$. After 5 epochs warming up, the learning rate is set to $10^{-2}$ for the first $230$ epochs, and decayed to $10^{-3}$ and $10^{-4}$ for another $20$ and $10$ epochs, respectively. Our method is implemented with the PyTorch library~\cite{paszke2017pytorch}.

{\noindent \textbf{Inference}.} During the inference phase, the STC first filters the anchors on the first three feature maps with the positive confidence scores smaller than the threshold $\theta=0.01$, and then the STR adjusts the anchors on the last three feature maps. The second step keeps top $2000$ high detections among these refined anchors. Finally, we apply the non-maximum suppression (NMS) with jaccard overlap of $0.4$ to generate $750$ high confident results per image. The multi-scale testing strategy is used during the inference phase.

\section{Result on WIDER FACE}
The WIDER FACE dataset contains $32,203$ images and $393,703$ annotated faces bounding boxes including high degree of variability in scale, pose, facial expression, occlusion and lighting condition. It is split into the training ($40\%$), validation ($10\%$) and testing ($50\%$) subsets by randomly sampling from each scene category (totally $61$ classes), and defines three levels of difficulty: Easy, Medium, Hard, based on the detection rate of EdgeBox~\cite{zitnick2014edge}. Following the evaluation protocol in WIDER FACE, we only train the model on the training set and test on both the validation and testing sets. To obtain the evaluation results on the testing set, we submit the detection results to the authors for evaluation.

\begin{figure*}[t]
\centering
\subfigure[Val: Easy]{
\label{fig:ve}
\includegraphics[width=0.45\linewidth]{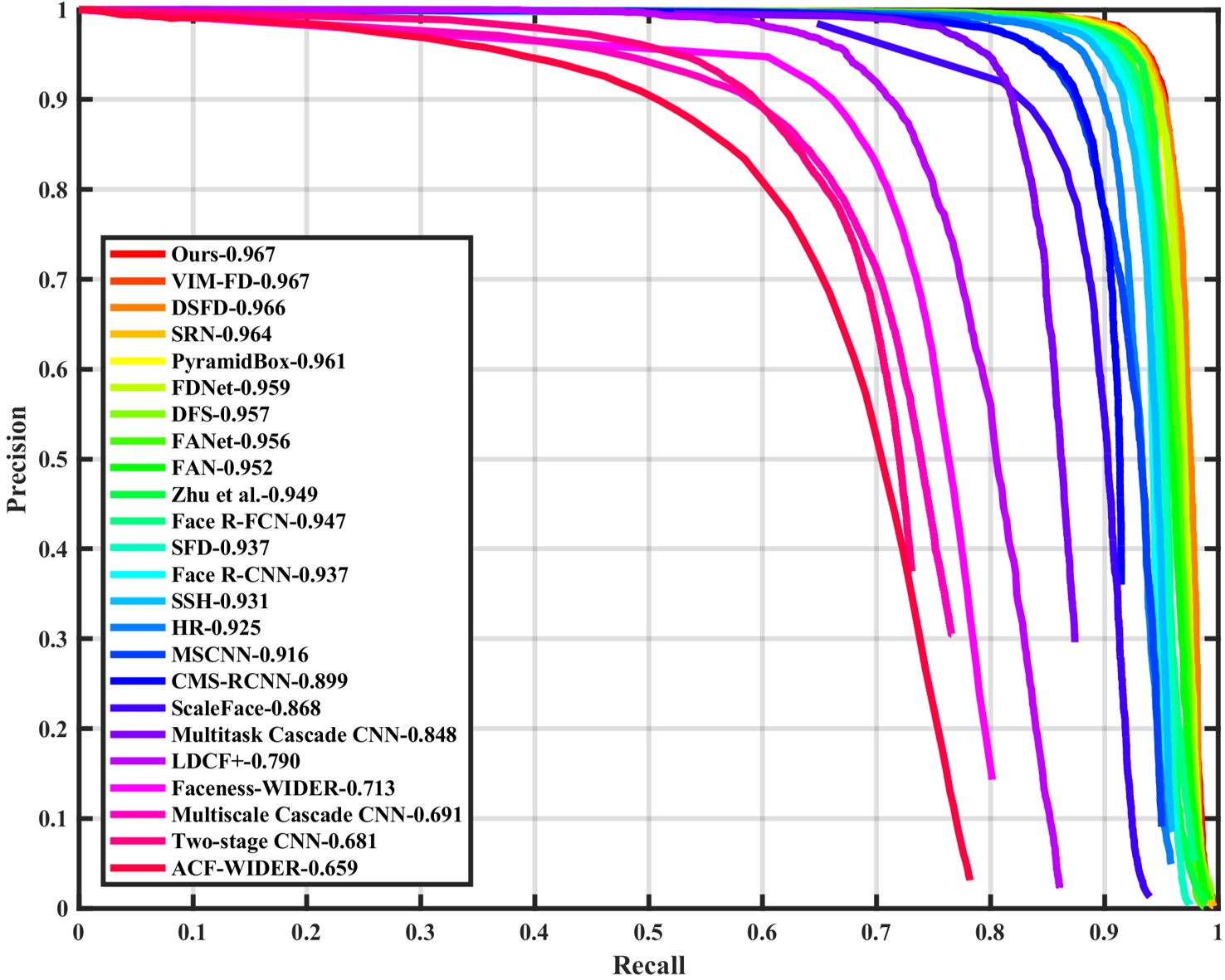}}
\subfigure[Test: Easy]{
\label{fig:te}
\includegraphics[width=0.45\linewidth]{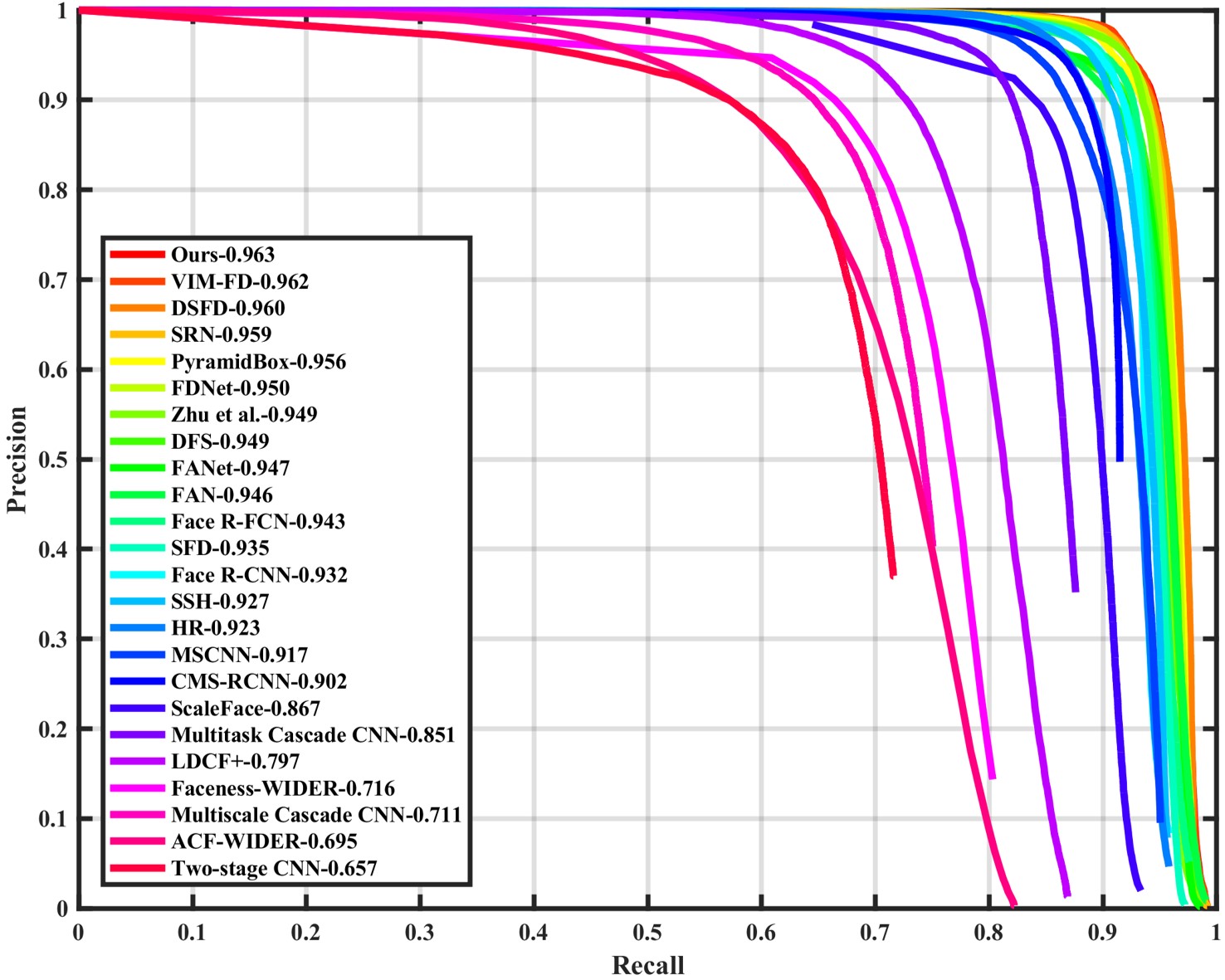}}
\subfigure[Val: Medium]{
\label{fig:vm}
\includegraphics[width=0.45\linewidth]{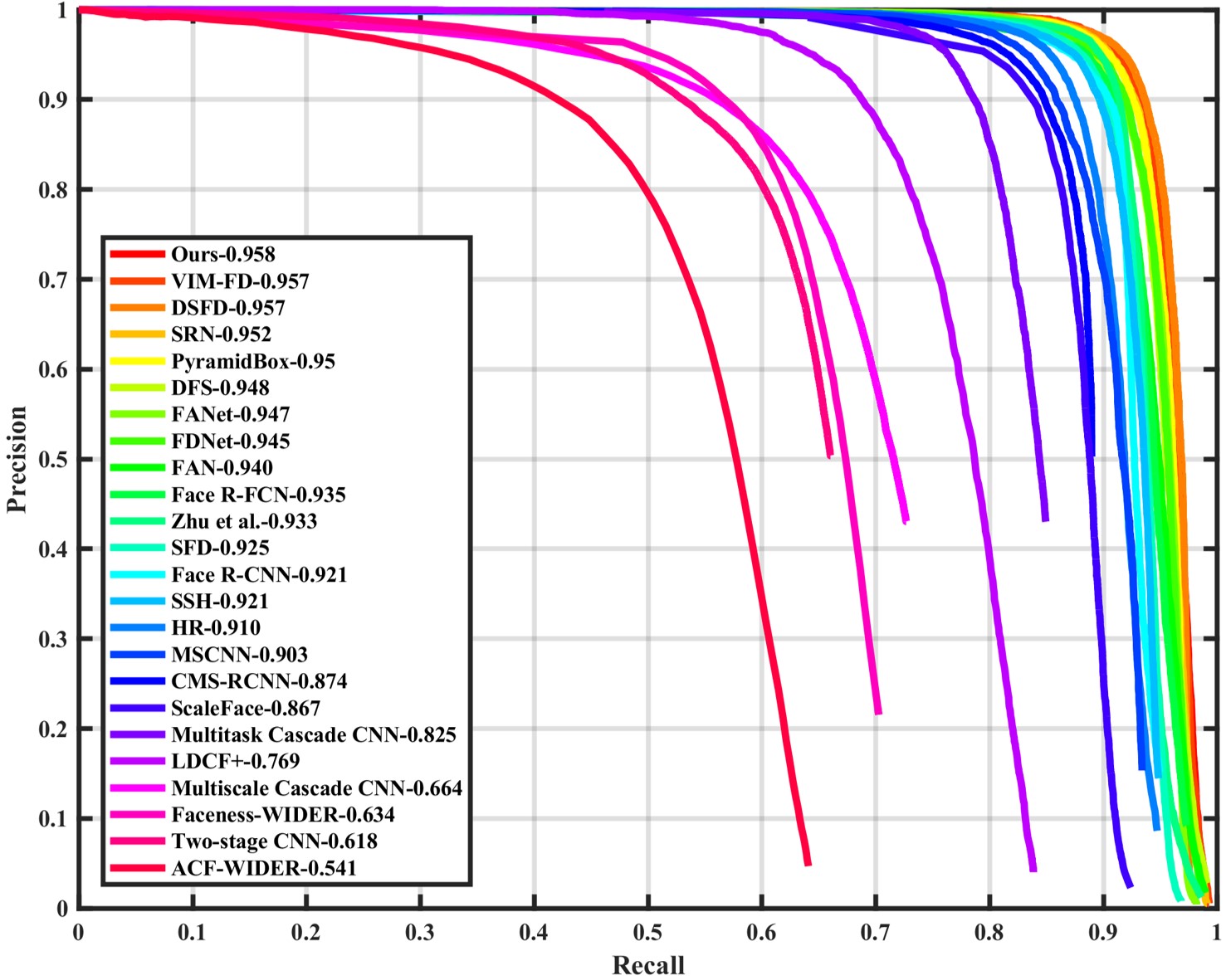}}
\subfigure[Test: Medium]{
\label{fig:tm}
\includegraphics[width=0.45\linewidth]{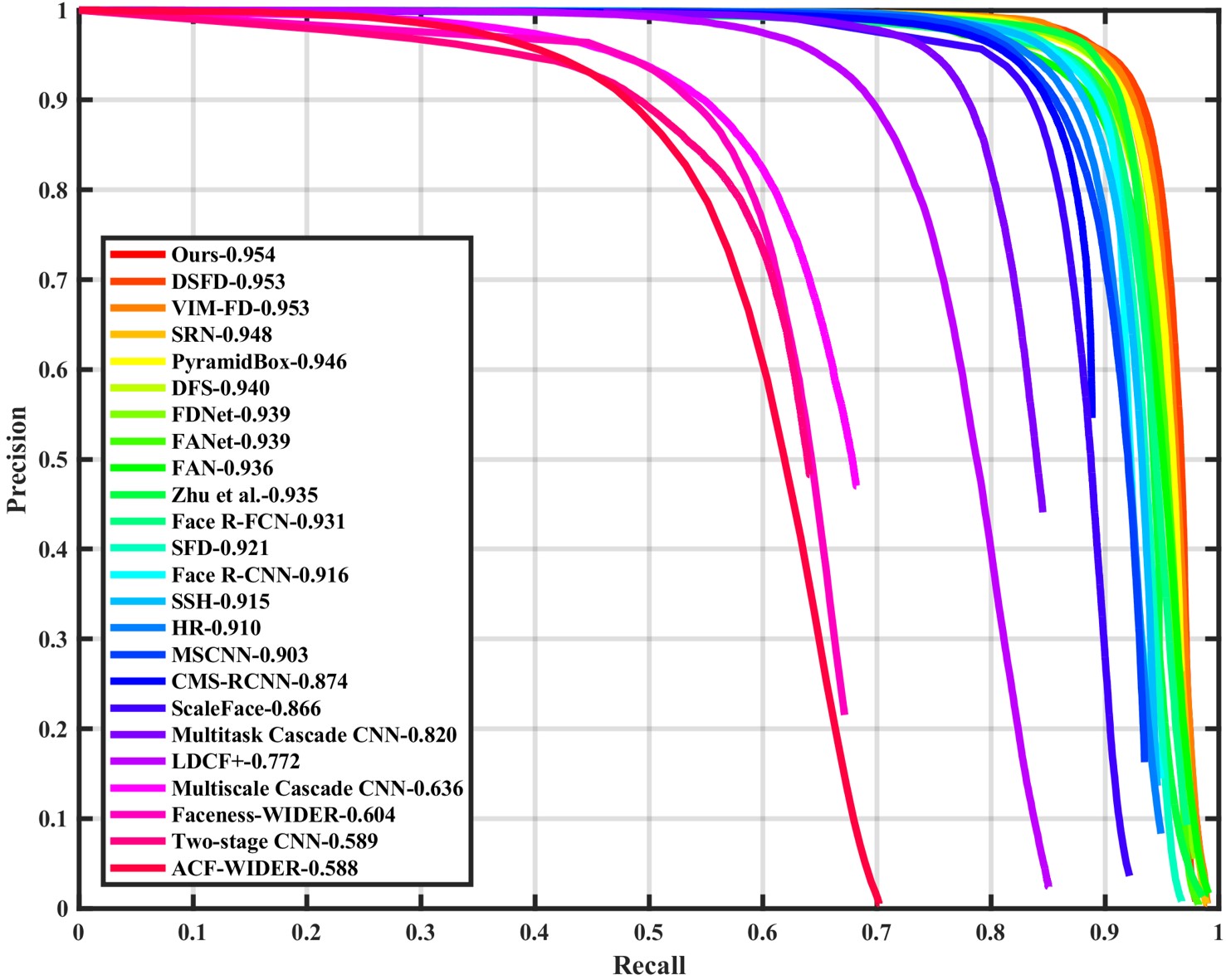}}
\subfigure[Val: Hard]{
\label{fig:vh}
\includegraphics[width=0.45\linewidth]{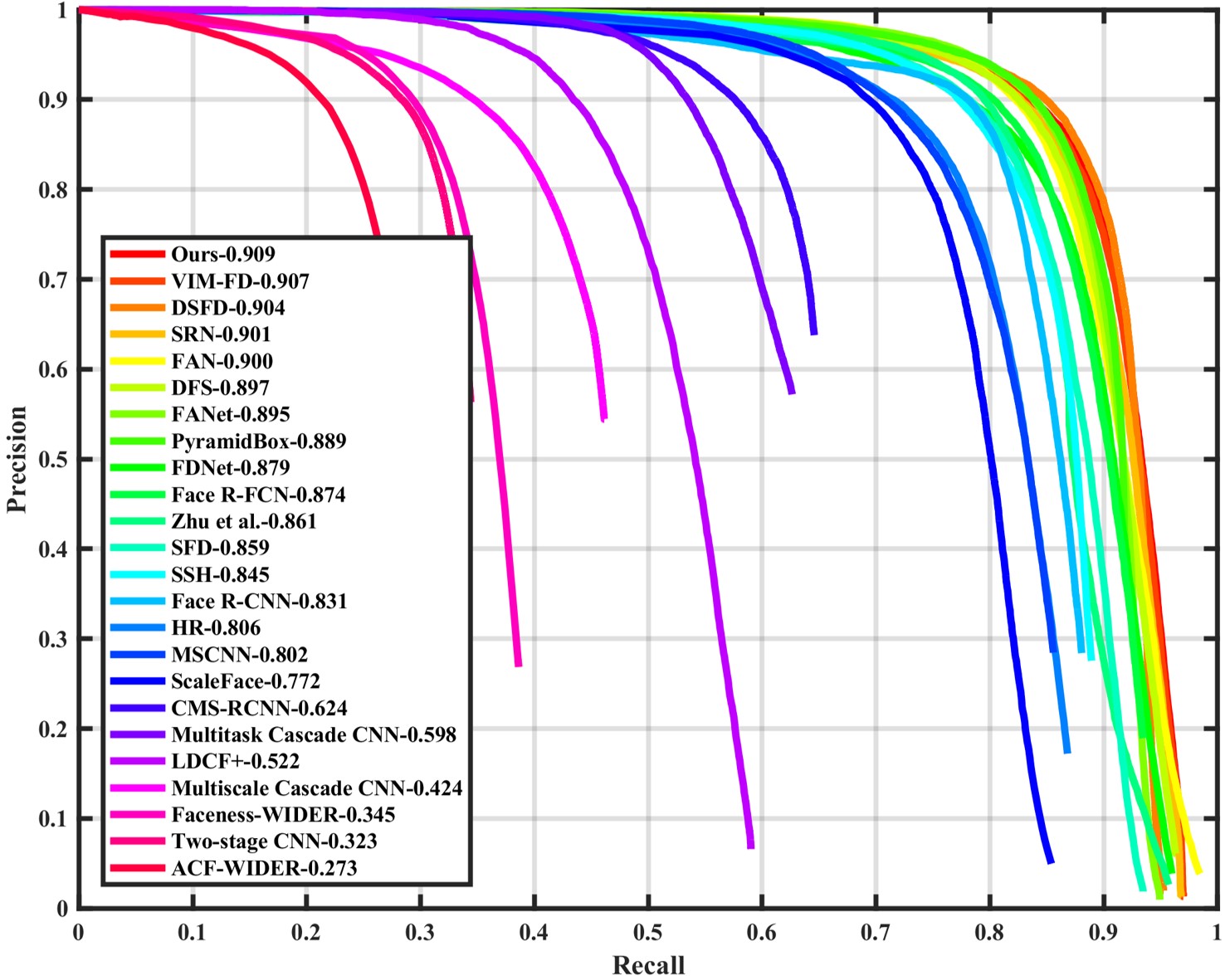}}
\subfigure[Test: Hard]{
\label{fig:th}
\includegraphics[width=0.45\linewidth]{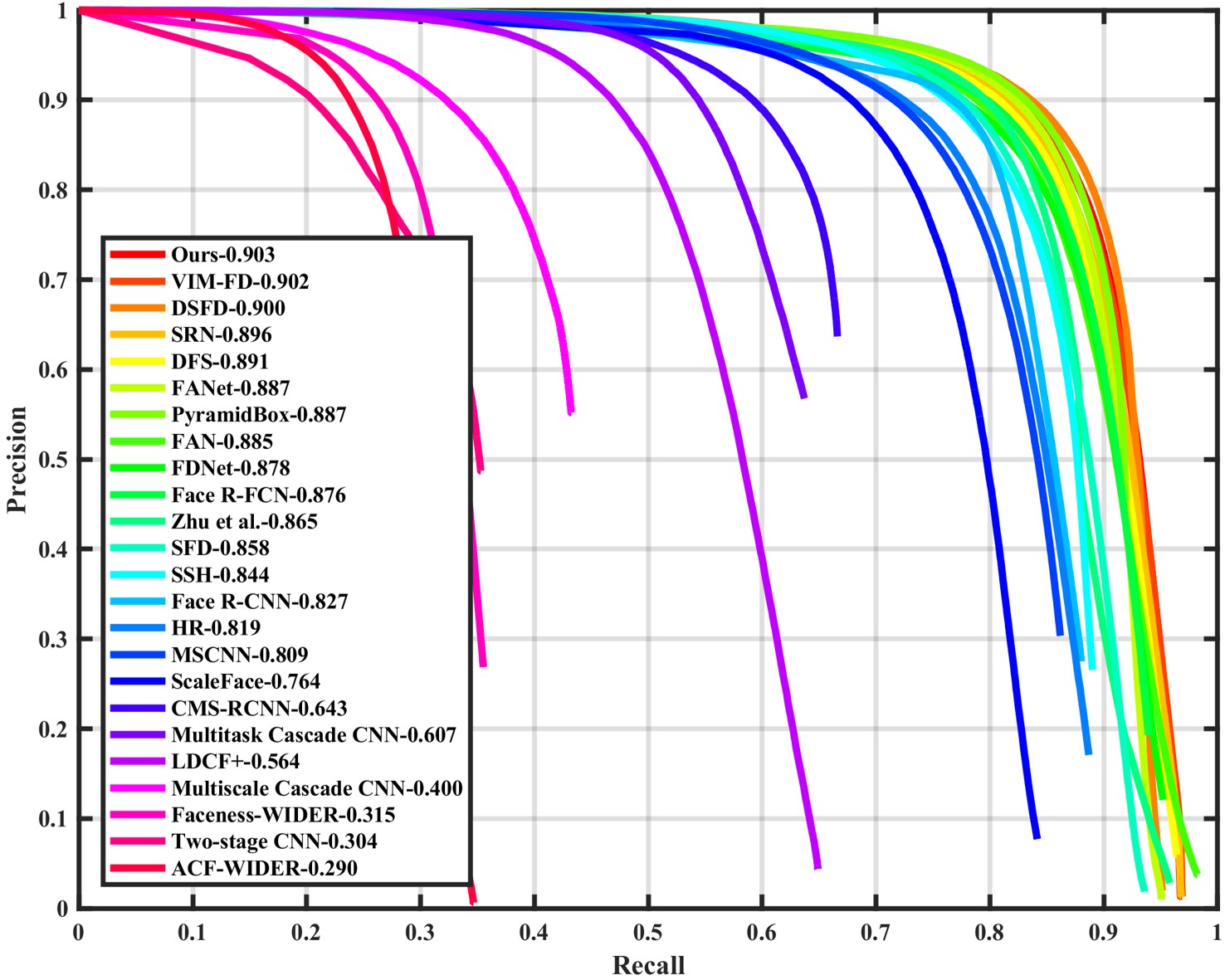}}
\caption{Precision-recall curves on WIDER FACE validation and testing subsets.}
\label{fig:wider-face-ap}
\end{figure*}

As shown in Figure~\ref{fig:wider-face-ap}, we compare our method (namely ISRN) with $23$ state-of-the-art face detection methods~\cite{yang2016wider, yang2014aggregate, yang2015facial, zhang2016joint, zhu2016cms, ohn2016boost, hu2017finding, wang2017face, wang2017detecting, yang2017face, najibi2017ssh, zhang2017s, cai2016unified, wang2017fan, zhu2018seeing, tang2018pyramidbox, zhang2018face, chi2018selective, zhang2017feature, li2018dsfd, tian2018learning, zhang2019robust}. We find that our model achieves the state-of-the-art performance  based on the average precision (AP) across the three evaluation metrics, especially on the Hard subset which contains a large amount of tiny faces. Specifically, it produces the best AP scores in all subsets of both validation and testing sets, \ie, $96.7\%$ (Easy), $95.8\%$ (Medium) and $90.9\%$ (Hard) for validation set, and $96.3\%$ (Easy), $95.4\%$ (Medium) and $90.3\%$ (Hard) for testing set, surpassing all approaches, which demonstrates the superiority of our face detector. We show one qualitative result of the World Largest Selfie in Figure~\ref{fig:slumia}. Our detector successfully finds about $900$ faces out of the reported $1,000$ faces.

\begin{figure*}
\centering
\includegraphics[width=0.9\linewidth]{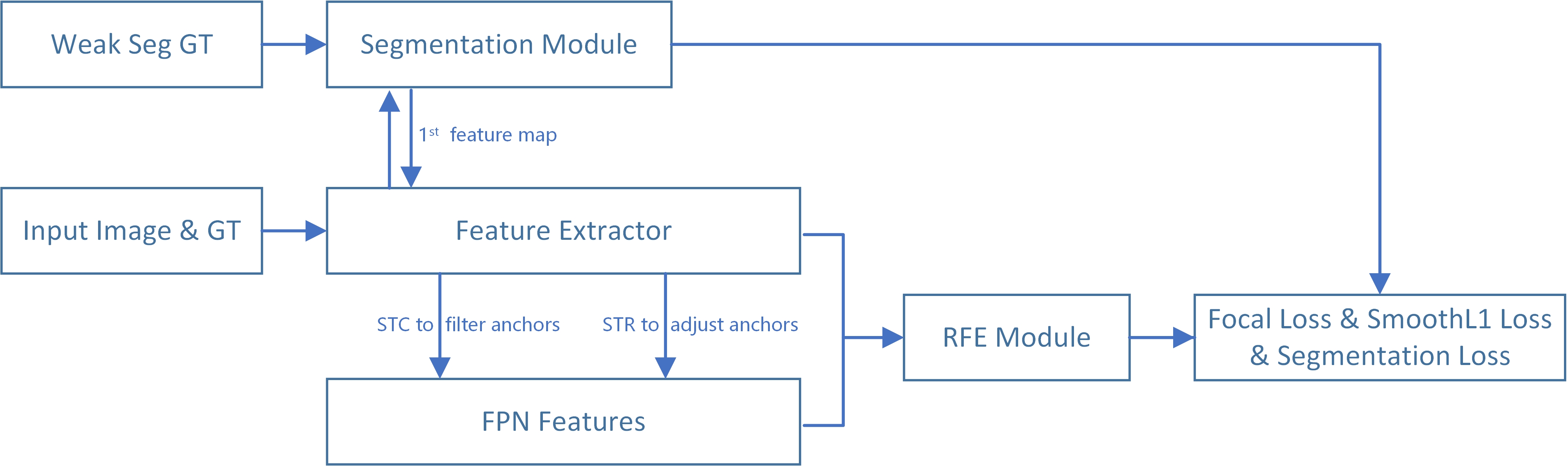}
\caption{The brief overview of Selective Refinement Network with segmentation branch.}
\label{fig:seg}
\end{figure*}

\section{Things We Tried That Did Not Work Well}
This section lists some techniques that do not work well in our model, probably because (1) we have a strong baseline, (2) combination of ideas is not trivial, (3) they are not robust enough for universality, and (4) our implementation is wrong. This does not mean that they are not applicable to other models or other datasets. 

{\noindent \textbf{Decoupled Classification Refinement (DCR)~\cite{cheng2018revisiting}.}}
It is an extra-stage classifier for the classification refinement. During the training process, DCR samples hard false positives with high condifence scores from the base Faster R-CNN detector, and then trains a stronger classifier. At the inference time, it simply multiplies the score from the base detector and another score from DCR to rerank detection results. Faster R-CNN with DCR gets great improvement on MS COCO~\cite{lin2014microsoft} and PASCAL VOC~\cite{everingham2010pascal} datasets. Therefore, we try to use DCR to suppress the false positives at the beginning and conduct some inquiring experiments based on our SRN baseline. However, with the help of RPN proposals and ROIs, the sampling strategies of DCR for two-stage detectors are much easier to design than the one for one-stage methods. SRN face detector produces too much boxes so we try a lot of sampling heuristics. Besides, we attempt some different crop size of the training examples due to the large scale variance and numerous small faces on WIDER FACE. Considering the scale of training set (positive and negative examples cropped from WIDER FACE training set) and network overfitting, we also try DCR backbones with different order of magnitude. With the setting of crop size=$20$, DRN-22 backbone and sampling strategy (positive examples: 0.5 $<$ IOU $<$ 0.8 and negative examples: IOU $<$ 0.3), our best result is also slight lower than our baseline detector. Further experiments about DCR need to be conducted for face detection task on WIDER FACE.

{\noindent \textbf{Segmentation Branch~\cite{zhang2018des}.}} A segmentation branch is added on SSD in DES, which is applied to the low level feature map and supervised by weak bounding-box level segmentation ground truth. The low level feature map is reweighed by the output $H\times W\times1$ map of segmentation branch. This enhancement can be regarded as a element-wise attention mechanism. As shown in Figure\ref{fig:seg}, we apply segmentation branch to the first low level feature map of SRN but the final results drop a bit on three metrics.

\begin{figure}[b]
\centering
\includegraphics[width=0.8\linewidth]{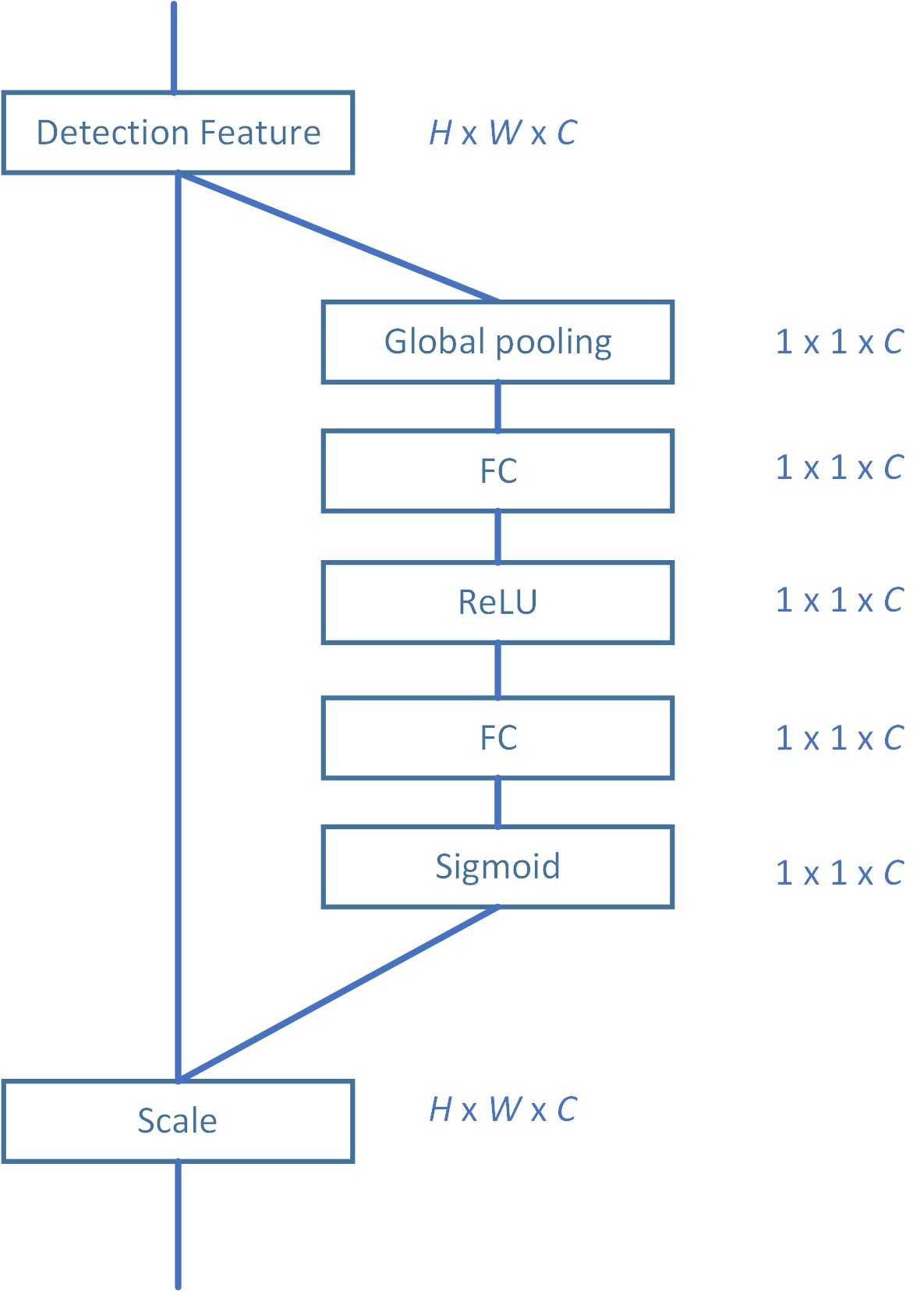}
\caption{Applying SE block to the detection feature.}
\label{fig:se}
\end{figure}

{\noindent \textbf{Squeeze-and-Excitation (SE) Block~\cite{hu2017squeeze}.}} It adaptively reweighs channel-wise features by using global information to selectively emphasise informative features and suppress useless ones. It can be regarded as a channel-wise attention mechanism with a squeeze-and-excitation $1 \times 1 \times C$ feature map, and the original feature will be reweighed to generate more representational one. As shown in Figure\ref{fig:se}, we apply SE block to the final detection feature map of SRN, but the final results drop $0.2\%$, $0.2\%$ and $0.4\%$ respectively on Easy, Medium and Hard metric.

\section{Conclusion}
To further boost the performance of SRN, we exploit some existing techniques including new data augmentation strategy, improved backbone network, MS COCO pretraining, decoupled classification module, segmentation branch and SE block. By conducting extensive experiments on the WIDER FACE dataset, we find that some of these techniques bring performance improvements, while few of them do not well adapt to our baseline. By combining these useful techniques together, we present an improved SRN detector  and obtain the state-of-the-art performance on the widely used face detection benchmark WIDER FACE dataset.

{\small
\bibliographystyle{ieee}
\bibliography{egbib}
}

\end{document}